\RequirePackage{fix-cm}

\documentclass[smallextended]{svjour3}       

\smartqed  
\usepackage{graphicx}

\usepackage{graphicx}
\usepackage{amsmath}
\usepackage{amssymb}

\usepackage{algorithmic}
\usepackage{algorithm}
\usepackage{subfigure}
\usepackage{array}
\usepackage{multirow,tabularx}

\newcommand{\bfx}{{\textbf{x}}}

\newcommand{\bfw}{{\textbf{w}}}

\begin{document}

\title{{Online classifier adaptation for cost-sensitive learning}}

\author{Junlin Zhang\and
Jos{\'e} Garc{\'i}a}

\institute{J. Zhang$^*$ \at
Xi'an University of Architecture and Technology, Xi'an, Shaanxi 710055, P.R. China\\
\email{junlinzhang1@yahoo.com}\\
$^*$J. Zhang is the corresponding author.
\and
J. Garc{\'i}a \at
Department of Computer Languages and Systems, Universitat Jaume I, Av. Sos Baynat s/n, 12071 Castel{\'i}o de la Plana, Spain\\
\email{josegarciajaumei@outlook.com}}

\maketitle

\begin{abstract}
In this paper, we propose the problem of online cost-sensitive classifier adaptation and the first algorithm to solve it. We assume we have a base classifier for a cost-sensitive classification problem, but it is trained with respect to a cost setting different to the desired one. Moreover, we also have some training data samples streaming to the algorithm one by one. The problem is to adapt the given base classifier to the desired cost setting using the steaming training samples online. To solve this problem, we propose to learn a new classifier by adding an adaptation function to the base classifier, and update the adaptation function parameter according to the streaming data samples. Given a input data sample and the cost of misclassifying it, we update the adaptation function parameter by minimizing cost weighted hinge loss and respecting previous learned parameter simultaneously. The proposed algorithm is compared to both online and off-line cost-sensitive algorithms on two cost-sensitive classification problems, and the experiments show that it not only outperforms them one classification performances, but also requires significantly less running time.
\end{abstract}

\section{Introduction}

In pattern recognition problems, we try to design a classification function to predict the class label of a data sample, so that the misclassification errors of a set of training samples can be minimized \cite{Bischl2012313,cherkassky2004practical,Dhanalakshmi2011350,Kusner20141939,Persello20132091,Charnay2013499}.
A popular assumption for the learning of a classifier is that the loss of misclassifying any data sample in the training set is equal. However, in real-world applications, different misclassifications may result in significant different costs. For example, in the problem of breast cancer diagnosis, misclassifying a malignant tumor sample may cause much more cost than misclassifying a benign tumor sample. Thus it is necessary to take the costs of different types of misclassifications into account when a classifier is trained. This problem is named cost-sensitive learning in machine learning community \cite{Tan19937,Greiner2002137,Zadrozny2003435,Elkan2001973,zhou2006training}.
Given the cost setting, i.e., costs of different misclassifications, the target of cost-sensitive learning is to train a classifier so that the cost of overall misclassification can be minimized.
In cost-sensitive binary classification, we can have different costs for misclassifications of positive and negative samples. In this case, misclassifying a positive sample to a negative sample incorrectly may results much higher cost than misclassifying a negative sample to a positive sample. So we must design a classify to correctly classify most of the positive samples, while allow some misclassification of negative samples. In this way, the overall misclassification cost can be minimized.
Lots of cost-sensitive learning algorithms have been proposed to take account of different misclassification costs. For example, Zhou et al. \cite{zhou2006training} proposed to
train cost-sensitive neural networks by using technologies of sampling and threshold-moving (STM), so that the distribution of the training data samples can be modify, and the costs of different types of misclassifications can be conveyed  by the appearances of the examples.
Sun et al. \cite{sun2007cost}  provided a comprehensive analysis of the AdaBoost algorithm regarding  its application  in the class imbalance problem, and  developed three cost-sensitive boosting algorithms (CSB), by introducing cost items into the learning framework of AdaBoost.
Masnadi-Shirazi and Vasconcelos \cite{masnadi2011cost} also proposed a AdaBoost-based cost-sensitive learning algorithm (ABC) to design cost-sensitive boosting algorithms,
by considering two necessary conditions for optimal cost-sensitive learning, which are the minimization of expected losses by optimal cost-sensitive decision rules, and the minimization of empirical loss to emphasize the neighborhood of the desired cost-sensitive boundary. Ting \cite{ting2002instance}  introduced a sample-weighting method (SW) to induce cost-sensitive trees, by generalizing the standard tree induction process and
initial instance weights determine the type of tree to be induced-minimum error trees or minimum high cost error trees. {Chen eta al. \cite{chen2014fast} proposed a supervised learning algorithm fast flux discriminant, for large-scale nonlinear cost-sensitive classification problems, by decomposing the kernel density estimation in the original feature space into selected low-dimensional subspaces. This method archives the efficiency, interpretability and accuracy simultaneously, and meanwhile it is also sparse and naturally handles mixed data types.}

With rapid development of internet technology, more and more data is generated continuously, and the training set of data samples is been increased every day with new data samples added to the training set. Moreover, the cost setting can also be changed from time to time. This proposed tow new challenges to the cost sensitive learning problems:

\begin{enumerate}
\item When the cost setting is changed, the learned classifier cannot be adapted to the new cost setting. A possible strategy to solve this problem is to learn a new classifier according to the new cost setting using the entire training set from the very beginning and ignores the previous learned classifier with previous cost setting. However this strategy is time-consuming, especially when the training set is large.

    When we already have a classifier learned according to a cost setting, can we utilize it to learn another classifier with regard to a different cost setting? This problem is defined as classifier adaption.
    Actually, classifier adaptation has been applied to performance measures optimization \cite{Li20131370} and cross-domain learning \cite{Yang2007188}. In \cite{Li20131370}, a classifier is learned to optimize a performance measure, and then adapted to optimize another performance measure, while in \cite{Yang2007188}, a classifier is learned from a domain, and then adapted to a different domain.  In this paper, we propose the problem of adapting a learned classifier to a different cost setting.

\item When the data samples are generated and added to the training set one by one, the transitional cost sensitive learning methods cannot be applied, since they assumes that the entire training set is given to the algorithm once.  {Recently,  cost-sensitive online classification (CSOC) method was proposed by Wang et al. \cite{Wang20121140}. This method takes the training set one by one and update the cost sensitive classifier online \cite{Patil2015349,Zaina2014114,Fonseca2007169}. However, CSOC is also constrained to fixed cost setting. When a cost setting is given, it learn a new classifier online, and ignores the other classifiers learned with different cost settings.}
    Can we learn a classifier from a base classifier trained with different cost sensitive setting online? This problem remains an open problem.
\end{enumerate}

To solve the above two problems simultaneously, in this paper, we propose the first online cost-sensitive classier adaption method. We assume that we have a existed cost-sensitive classifier, and we try to adapt it to another classifier with regard to a different cost setting, with help of data samples appearing one by one in an online way. The adaptation is implemented by adding an adaptation function, and the it is learned by updating the adaptation function parameter with the coming training samples with different misclassification costs. We construct an objective function by
considering the respecting previous learned and minimizing cost weighted hinge loss with coming training samples. By solving the objective function with a gradient descent method and we develop an iterative algorithm.
The contributions of this paper are of two folds:

\begin{enumerate}
\item We proposed the problem of online cost-sensitive classifier adaptation.
\item We proposed a novel algorithm to solve this problem.
\end{enumerate}
The rest parts of this paper are organized as follows: in Section \ref{sec:method}, we introduce the proposed novel method. In Section \ref{sec:experiment}, the proposed method is evaluated on some benchmark data sets. In Section \ref{sec:conclusion}, the paper is concluded with some future works.

\section{Proposed Method}
\label{sec:method}

\subsection{Problem Formulation}

In this paper, instead of learning a novel cost-sensitive classifier from the given training set and the cost setting,
we hope to use the existed classifier by employing the framework of classifier adaptation to learn the cost-sensitive classifier effectively. Suppose that we already have a classifier $f_0(\bfx)$ learned without consider the different costs of misclassifications of positive and negative samples, or a classifier learned with different cost setting, we want to adapt it to a problem with a new cost setting. To this end, we construct a new classifier $f(\bfx)$ by adding a  linear adaptation function $\bfw^\top \bfx$ to $f_0(\bfx)$, i.e.

\begin{equation}
\begin{aligned}
f(\bfx)=f_0(\bfx)+\bfw^\top \bfx
\end{aligned}
\end{equation}
where $\bfw \in \mathbb{R}^d$ is the adaptation function parameter. Please note that $f_0(\bfx)$ can be any type of classifier, for example, SVM, Adaboost, etc. In this way, we transfer the problem of cost-sensitive classifier adaptation to the learning of $\bfw$.

In the traditional cost-sensitive learning problem, a training data set composed of many positive and negative training samples are given. The cost factors of misclassification of positive and negative samples are denoted as $C_+$ and $C_-$ respectively. Please note that when we train $f_0(\bfx)$, $C_+$ and $C_-$ are set to different values.
The target of cost-sensitive learning is to learn a classifier which could minimize the overall cost of misclassification of the training samples. However, in the online learning scene, we do not have the entire training data set during the training procedure. Instead, the training data samples are given sequentially, and the algorithm is run in an iterative way. In each iteration, only one training sample is given, and the classifier is updated only with regard to this training sample. In the $t$-th iteration, we assume that wa have a training sample $(\bfx_t,y_t)$, where $\bfx_t\in \mathbb{R}^d$ is its $d$-dimensional feature vector, and $y_t\in \{+1,-1\}$ is its corresponding class label. The corresponding misclassification cost is also given as $C_t$,

\begin{equation}
C_t =
\left \{
\begin{aligned}
C_+,&if~y_i=+1;\\
C_-,&if~y_i=-1.
\end{aligned}
\right .
\end{equation}
Moreover, we also assume that we already learned an adaptation function parameter from the previous iteration $\bfw_{t-1}$. To update $\bfw$, we consider the following two problems.

\begin{itemize}
\item \textbf{Respecting previous learned $\bfw_{t-1}$}: To make the learned $\bfw$ consistent, we hope the updated $\bfw_t$ to respect the previous $\bfw_{t-1}$. To this end, we minimize the squared $\ell_2$ distance between them,

\begin{equation}
\label{equ:obj1}
\begin{aligned}
\underset{\bfw}{\min}
~&
\frac{1}{2} \left \| \bfw-\bfw_{t-1}\right\|_2^2.
\end{aligned}
\end{equation}

\item \textbf{Minimizing Cost Weighted Hinge Loss}: To measure the loss of misclassification, we apply the hinge loss function to $(\bfx_t, y_t)$, which is defined as

\begin{equation}
\begin{aligned}
L(y_t, f(\bfx_t))
&=\max(0,1-y_t f(\bfx_t))\\
&=\max
\left (0, 1- y_t \left (f_0(\bfx_t)+\bfw^\top \bfx_t \right)\right).
\end{aligned}
\end{equation}
Since positive and negative samples have different misclassification costs, we weight the hinge loss of the $t$-th sample by its corresponding cost factor $C_t$, and minimize the weighted loss,

\begin{equation}
\begin{aligned}
\underset{\bfw}{\min} ~
 C_t \times \max(0,1- y_t \left (f_0(\bfx_t)+\bfw^\top \bfx_t \right))
\end{aligned}
\end{equation}
By introducing a nonnegative slack variable $\xi$, the optimization problem with a cost weighted hinge loss is transferred to

\begin{equation}
\label{equ:obj2}
\begin{aligned}
\underset{\bfw,\xi}{\min}
~&
C_t \xi,\\
s.t.
~&
1- y_t \left (f_0(\bfx_t)+\bfw^\top \bfx_t \right) \leq \xi, 0 \leq \xi.
\end{aligned}
\end{equation}
Here the cost factor $C_t$ is similar to the penalty factor of SVM. However, we must note that this penalty factor is cost-sensitive.

\end{itemize}

By considering the problems in (\ref{equ:obj1}) and (\ref{equ:obj2}) simultaneously, we obtain the optimization problem for the updating of $\bfw$ in the $t$-th iteration,

\begin{equation}
\label{equ:obj3}
\begin{aligned}
(\bfw_{t},\xi_t)=\underset{\bfw,\xi}{\arg\min}
~&
\frac{1}{2} \left \| \bfw-\bfw_{t-1}\right\|_2^2 + \alpha C_t \xi,\\
s.t.
~&
1- y_t \left (f_0(\bfx_t)+\bfw^\top \bfx_t \right) \leq \xi, 0 \leq \xi.
\end{aligned}
\end{equation}
where $\alpha$ is a tradeoff parameter, and it is chosen by {cross-fold validation on a training set}.
By solving this problem, we can obtain an adaptation function parameter $\bfw_t$ with regard to the training sample input in the $t$-th iteration. We should note that the obtained $\bfw_t$ is learned from both the previous $\bfw_{t-1}$ and the training sample $(\bfx_t,y_t)$. Most importantly, the updating of $\bfw_t$ relies on the cost of misclassification cost of $(\bfx_t,y_t)$ by considering the cost factor as a loss weight. When different samples come, different cost factor is used and the hinge loss is weighted correspondingly.

\subsection{Optimization}

To optimize the objective function in (\ref{equ:obj3}), we use the Lagrange multiplier method. The Lagrange function is
\begin{equation}
\label{equ:Lag0}
\begin{aligned}
\mathcal{L}(\bfw,\xi,\tau,\lambda)=
&
\frac{1}{2} \left \| \bfw-\bfw_{t-1}\right\|_2^2 + \alpha C_t \xi\\
&+
\tau \left ( 1- y_t \left (f_0(\bfx_t)+\bfw^\top \bfx_t \right) - \xi \right )
- \lambda \xi.
\end{aligned}
\end{equation}
where $\tau$ is the nonnegative Lagrange multiplier for the constrain of $1- y_t \left (f_0(\bfx_t)+\bfw^\top \bfx_t \right) \leq \xi$, and $\lambda$ is the nonnegative Lagrange multiplier for the constrain of $0 \leq \xi$.
According to the dual theory of optimization, the minimization of (\ref{equ:obj3}) can be achieved by solving the following dual problem,

\begin{equation}
\label{equ:object1}
\begin{aligned}
\max_{\tau,\lambda} \min_{\bfw,\xi}
~
&
\mathcal{L}(\bfw,\xi,\tau,\lambda)\\
s.t.~&
\tau \geq 0, \lambda \geq 0.
\end{aligned}
\end{equation}
To solve this problem, we set the divertive of the Lagrange function $\mathcal{L}(\bfw,\xi,\tau,\lambda)$ with regard to $\bfw$ to zero, and we have

\begin{equation}
\label{equ:Lag1}
\begin{aligned}
\frac{\partial \mathcal{L}}{\partial \bfw}=
&
\left ( \bfw-\bfw_{t-1}\right) -
\tau  y_t  \bfx_t =0
\Rightarrow  \bfw = \bfw_{t-1} + \tau  y_t  \bfx_t.
\end{aligned}
\end{equation}
{\textbf{Remark}: The motivation of setting the divertive of the Lagrange function with regard to $\bfw$ to zero is to solve the inner minimization problem. In equation (\ref{equ:object1}), the problem is coupled with two optimization problems, which is a inner minimization problem, and an outer maximization problem. To solve this coupled problem, our strategy is first solving the inner problem, and then substituting the results to the objective to solve the outer problem. According to the optimization theory, the minimization of $\mathcal{L}$ with regard to $\bfw$ is reached at a solution making its divertive zero, thus we should set the divertive of the $\mathcal{L}$ with regard to $\bfw$ to zero to obtain the optimal $\bfw$.}

Moreover, we also set its divertive with regard to $\xi$ to zero, and obtain

\begin{equation}
\label{equ:Lag2}
\begin{aligned}
\frac{\partial \mathcal{L}}{\partial \xi}=
&\alpha C_t - \tau - \lambda = 0
\Rightarrow
\alpha C_t - \tau = \lambda  \geq 0
\Rightarrow
\tau \leq \alpha C_t.
\end{aligned}
\end{equation}
Substituting results of both (\ref{equ:Lag1}) and (\ref{equ:Lag2}) to the Lagrange function in (\ref{equ:Lag0}), we can rewrite it as the function of only variable $\tau$,

\begin{equation}
\begin{aligned}
\mathcal{L}(\tau)=
&
\frac{1}{2} \left \| \tau  y_t  \bfx_t  \right\|_2^2 +
\tau \left [ 1- y_t \left (f_0(\bfx_t)+\left(\bfw_{t-1} + \tau  y_t  \bfx_t \right)^\top \bfx_t \right)  \right ]\\
=&
\frac{1}{2} \tau^2 \bfx_t^\top \bfx_t +
\tau \left [ 1- y_t \left (f_0(\bfx_t)+\bfw_{t-1}^\top \bfx_t \right ) \right ] - \tau^2    \bfx_t^\top \bfx_t  \\
=&
-\frac{1}{2} \tau^2 \bfx_t^\top \bfx_t +
\tau \left [ 1- y_t \left (f_0(\bfx_t)+\bfw_{t-1}^\top \bfx_t \right ) \right ].
\end{aligned}
\end{equation}
By setting the divertive of $\mathcal{L}(\tau)$ with regard to $\tau$ to zero, we have the initial solution of $\tau$,

\begin{equation}
\begin{aligned}
&\frac{\partial \mathcal{L}(\tau)}{\partial \tau}
=
- \tau \bfx_t^\top \bfx_t +
 \left [ 1- y_t \left (f_0(\bfx_t)+\bfw_{t-1}^\top \bfx_t \right ) \right ] = 0 \\
&\Rightarrow
\tau = \frac{ 1- y_t \left (f_0(\bfx_t)+\bfw_{t-1}^\top \bfx_t \right ) }{\bfx_t^\top \bfx_t}.
\end{aligned}
\end{equation}
Moreover, we should also note that in (\ref{equ:Lag0}), we have a constrain $\tau \geq 0$, and in (\ref{equ:Lag2}) we have another constrain $\tau \leq \alpha C_t$. Thus the solution of $\tau$ must fall in the following range,

\begin{equation}
\begin{aligned}
0\leq \tau \leq \alpha C_t.
\end{aligned}
\end{equation}
In this way, the solution of $\tau^t$ can be obtained by discussing the following three cases:

\begin{enumerate}
\item \textbf{Case I}: When
$\frac{1- y_t \left (f_0(\bfx_t)+\bfw_{t-1}^\top \bfx_t \right ) }{\bfx_t^\top \bfx_t} \leq 0$, the solution of $\tau_t$ is

\begin{equation}
\label{equ:tau1}
\begin{aligned}
\tau_t=  0,
\end{aligned}
\end{equation}
 so that the constrain $\tau \geq 0$ can be satisfied.
\item \textbf{Case II}: When
$0 <\frac{1- y_t \left (f_0(\bfx_t)+\bfw_{t-1}^\top \bfx_t \right ) }{\bfx_t^\top \bfx_t} \leq \alpha C_t$, the solution of $\tau_t$ is

\begin{equation}
\begin{aligned}
\tau_t = \frac{1- y_t \left (f_0(\bfx_t)+\bfw_{t-1}^\top \bfx_t \right ) }{\bfx_t^\top \bfx_t},
\end{aligned}
\end{equation}
so that the minimization of (\ref{equ:obj3}) can be archived.

\item \textbf{Case III}: When
$  \alpha C_t< \frac{1- y_t \left (f_0(\bfx_t)+\bfw_{t-1}^\top \bfx_t \right ) }{\bfx_t^\top \bfx_t}$, we have the solution of $\tau_t$ as

\begin{equation}
\label{equ:tau3}
\begin{aligned}
\tau_t=  \alpha C_t,
\end{aligned}
\end{equation}
 so that the constrain $\tau_t \leq  \alpha C_t$ can be satisfied.
\end{enumerate}
After $\tau_t$ is determined, we can then update $\bfw_t$ using the result in (\ref{equ:Lag1}) as follows,

\begin{equation}
\label{equ:w}
\begin{aligned}
\bfw_t = \bfw_{t-1} + \tau_t  y_t  \bfx_t.
\end{aligned}
\end{equation}
It could be note that the new classifier adaptation function parameter is obtained by adding a bias term $y_t \bfx_t$ determined by the $t$-th sample to the previous $\bfw_{t-1}$. The bias term is weighted by a Lagrange multiplier $\tau_t$ which is further controlled by the cost factor of the $t$-th sample.

\subsection{Algorithm}

Based on the optimization results, we can develop an online cost-sensitive classifier adaptation algorithm which can take training samples one by one.
The algorithm takes an initial classifier $f(\bfx)$ as an input, and operates on a iterative way.
In each iteration, one new training sample is input to update the classifier adaptation function parameter, based on the updating rules in (\ref{equ:tau1}) - (\ref{equ:tau3}), and (\ref{equ:w}).

\begin{algorithm}[htb!]
\caption{Online Cost-Sensitive Classifier Adaptation algorithm (OCSCA).}
\label{alg:OCSCA}
\begin{algorithmic}
\STATE \textbf{Input}: An initial classifier function $f_0(\bfx)$;
\STATE {\textbf{Input}: Tradeoff parameter $\alpha$};

\STATE Initialize $t=0$ and $\bfw_0 = \textbf{0}$

\WHILE{A new training sample $(\bfx_t,y_t)$ with its corresponding misclassification cost $C_t$ is input}

\STATE Compute the initial solution of Lagrange multiplier $\tau$ as

\begin{equation}
\begin{aligned}
\tau_t' = \frac{ 1- y_t \left (f_0(\bfx_t)+\bfw_{t-1}^\top \bfx_t \right ) }{\bfx_t^\top \bfx_t}.
\end{aligned}
\end{equation}

\STATE Update $\tau_t$ as

\begin{equation}
\tau_t=
\left \{
\begin{aligned}
0,  & if~ \tau_t' \leq 0,\\
\tau_t', & if~ 0< \tau_t' \leq \alpha C_t\\
\alpha C_t, & if ~\tau_t' > \alpha C_t.
\end{aligned}
\right .
\end{equation}

\STATE Update $\bfw_t$ as

\begin{equation}
\begin{aligned}
\bfw_t = \bfw_{t-1} + \tau_t  y_t  \bfx_t.
\end{aligned}
\end{equation}

\STATE Update $t=t+1$;

\ENDWHILE

\STATE \textbf{Output}:
Output the learned cost-sensitive classifier function $f(\bfx)=f_0(\bfx)+{\bfw_{t-1}}^\top \bfx$
\end{algorithmic}
\end{algorithm}

\section{Experiments}
\label{sec:experiment}

In this section, we studied the proposed algorithm experimentally.

\subsection{Data sets}

In the experiments, we used two cost-sensitive learning data sets, which are introduced as follows.

\subsubsection{Face detection data set}

The first data set is a face detection data set used in \cite{viola2004robust}. This data set is a large data set, and it contains 9832 face images and 9832 non-face images. Each face image is treated as a positive sample, while each non-face image is treated as a negative sample. Moreover, each image is represented as  50,000 dimensional visual feature vector. The problem of face detection is to classify a given candidate image to face or non-face.  Moreover, we set the cost of misclassifying a face to non-face as 5, and that of misclassifying a non-face to fact to 1.

\subsubsection{Car detection data set}

The second data set we used is a car detection data set \cite{agarwal2004learning}. This data set contains 500 car images and 500 non-car images. The problem of car detection is to classify a given candidate image to car or non-car. In this problem car images are defined as positive images, and the non-car images are defined as negative images. In this case, we set the cost of misclassification of a car image to 8, and that of a non-car image to 1.

\subsection{Experiment setup}

To conduct the experiment, we used the 10-fold cross validation. An entire data set was split into 10 folds randomly, and then each set is used as a test set, while the remaining 9 sets were combined as a training set. Moreover, since the proposed method is based on the adaption of a classifier $f_0(\bfx)$ trained with different cost setting, we further split the training set to two subsets. The first subset contains 2 folds, and we used it to train $f_0(\bfx)$ with different cost setting. For the first data set, we used the cost setting of $C_+=2$ and $C_-=1$ to train $f_0(\bfx)$, and for the second data set, used $C_+=3$ and $C_-=1$. The second subset contains 7 folds, and we used it to learn $\bfw$ using the proposed online learning algorithm, by inputting the training samples of the second training subset to the algorithm one by one.

The classification performances were measured by the average classification accuracies and the average misclassification costs. They are defined as follows,

\begin{equation}
\begin{aligned}
&Average ~classification~accuracy = \frac{\sum_{i:\bfx_in\in \mathcal{T}}I(y_i = y_i^*)}{\sum_{i:\bfx_i\in \mathcal{T}} 1},\\
&Average~misclassification~cost=\frac{\sum_{i:\bfx_in\in \mathcal{T}}C_iI(y_i \neq y_i^*)}{\sum_{i:\bfx_i\in \mathcal{T}} 1},
\end{aligned}
\end{equation}
where $\mathcal{T}$ is the test set, $y_i^*$ is the predicted class label, and $I(y_i = y_i^*)=1$ if $y_i = y_i^*$, and 0 otherwise.

\subsection{Results}

We first compared the purposed online cost sensitive learning algorithm based on classifier adaptation to an online cost sensitive learning algorithm without considering the existed classifier $f_0(\bfx)$, and then compared it to some transitional cost sensitive learning algorithm. Because the proposed algorithm is the only method that can take advantage of $f_0(\bfx)$, for fear comparison, when we used the other algorithm, both $f_0(\bfx)$ and the 2 folds in the training set used to train $f_0(\bfx)$ were ignored.

\subsubsection{Comparison to online cost sensitivity classification method}

\begin{figure}[!htb]
  \centering
  \subfigure[Face detection data set]{
  \includegraphics[width=\textwidth]{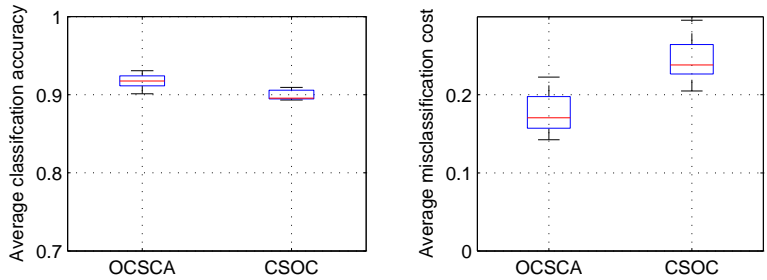}}
  \subfigure[Car detection data set]{
  \includegraphics[width=\textwidth]{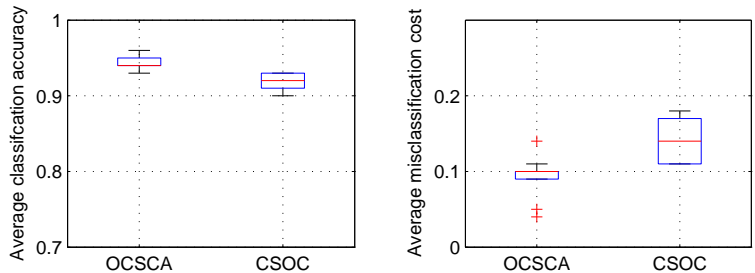}}
  \\
  \caption{Classification accuracies and misclassification costs of two online cost learning algorithms.}
  \label{fig:online}
\end{figure}

The boxplots of the classification accuracies and misclassification of the proposed  online cost-sensitive classifier adaptation algorithm (OCSCA) and the CSOC algorithm over 10-fold cross validation are given in Fig. \ref{fig:online}. From this figure, we can see that the proposed method outperforms the CSOC algorithm on both average accuracy and misclassification cost. Especially in the case of misclassification cost, the proposed algorithm achieves completely lower average misclassification cost then CSOC. This is because the proposed method takes advantage of an existing predictor learned from more data points by adapting it to a given cost setting. Even the existing predictor is learned according to a different cost setting. This is an strong evidence of the fact that classifier adaptation can benefit cost sensitive learning.

\subsubsection{Comparison to off-line cost sensitivity classification method}

\begin{figure}[!htb]
  \centering
  \subfigure[Face detection data set]{
  \includegraphics[width=\textwidth]{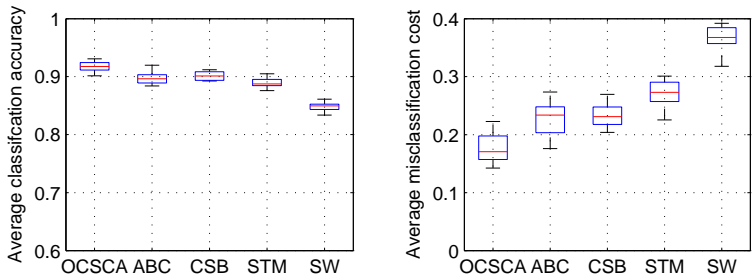}}\\
  \subfigure[Car detection data set]{
  \includegraphics[width=\textwidth]{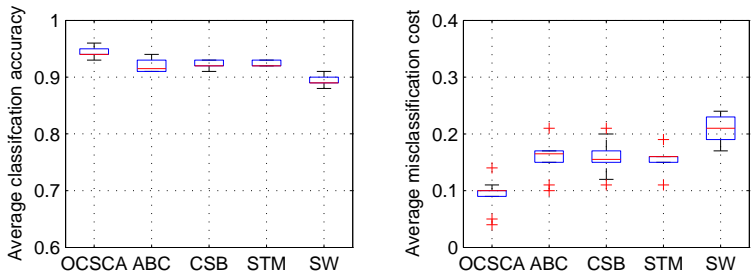}}\\
  \caption{Classification accuracies and misclassification costs of the proposed algorithm and off-line cost-sensitive learning algorithms.}\label{fig:offline}
\end{figure}

We also compare the proposed OCSCA to four most popular off-line cost-sensitive learning algorithms, which are STM proposed by Zhou et al. \cite{zhou2006training}, CSB proposed by Sun et al. \cite{sun2007cost}, ABC proposed by Masnadi-Shirazi and Vasconcelos \cite{masnadi2011cost}, and SW proposed by Ting \cite{ting2002instance}.  The boxplots the classification accuracies and misclassification costs are given in Fig. \ref{fig:offline}.  It is clear that in both the two figures, the proposed algorithm outperforms the compared algorithms on both  classification accuracies and misclassification costs. The outperforming is even more significant on the misclassification costs. A main reason for this phenomenon lies on the fact that the proposed OCSCA algorithm starts learning from a base classifier $f_0(\bfx)$, and then adapt it to the given cost setting via a training set, while the rest algorithms ignores $f_0(\bfx)$ and directly learn the classifier from the training set. This means using a base classifier and adapting it to a training set can significantly boost the performance of cost-sensitive learning. Moreover, among the compared algorithms, it seems ABC and CSB performs slightly better than the other two ones. A possible reason is that they use the formula of Adaboost algorithm \cite{Riccardi20141898,Mei2014,Ahachad2014,Chen2015277}, which performs well on detection problems.

\subsubsection{Running time}

\begin{figure}[!htb]
  \centering
    \includegraphics[width=0.7\textwidth]{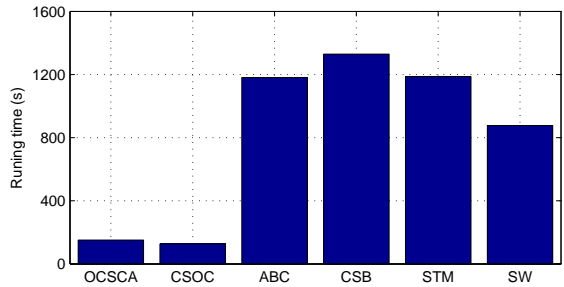}\\
  \caption{Running time of learning procedure of online learning algorithms and off-line algorithms.}
  \label{fig:time}
\end{figure}

An important advantage of the proposed OCSCA algorithm is its low time complexity compared to off-line algorithms. Thus we also compared the running time of these methods and the results are given in Fig. \ref{fig:time}. It is obverse that the running time of the  two online learning algorithms OCSCA and CSOC  is much less than that of the off-line learning algorithms. Both of OCSCA and CSOC take less than 200 seconds, while all the off-line learning algorithms take more than 800 seconds. This is not surprising because in each iteration, OCSCA and CSOC update the classifier using only one data sample, while the off-line learning algorithms needs to consider all the training samples.

\section{Conclusions and future works}
\label{sec:conclusion}

In this paper, we propose the problem of adapting an existing base classifier to a cost-sensitive classification problem. The base classifier is trained using different cost settings. Moreover, we proposed a novel online learning algorithm for the adaptation of the classifier. The algorithm takes one data sample at one time to update the adaptation parameter. The advantages of this method are of two folds:

\begin{enumerate}
  \item It can use the base classifier to boost the classification performance, and
  \item its running time is low due to its online learning nature.
\end{enumerate}

In this work, we used the SVM as the formulation of learning. In the future, we will study other classification methods, such as Adaboost. {We will design an iterative algorithm to learn the classifier online by adapting an existing classifier trained with a different cost setting, and the adaptation function is an combination of some candidate weak classifiers. In each iteration, we have select a weak classifier according to the classification cost of the coming data point, and update its weight. Moreover, the loss function of Adaboost will be modified to consider the classification costs.}
Moreover, we will also investigate the application of the proposed algorithm to information security \cite{xu2014evasion,xu2012stochastic,zhan2013characterizing,xu2014adaptive,xu2014evasion,zhan14characterization,Measuring}, bioinformatics \cite{wang2014computational,liu2013structure,zhou2014biomarker}, medial imaging \cite{Zhao3483264,7024999,zhao2013}, computer vision \cite{6984567,Li2011885,Li2010242,wan2012efficient,wan2011topology,vTRUST,TBaaS,wang2013can,wang2013gender,wang2013hete,guo2012kinship}, reinforcement learning \cite{IAT08,GEC09}, cloud computing \cite{eScience13,ScienceCloud14}
and microprocessor reliability modeling \cite{6915733,chen2013impact,chen2014simulation,chen2012backend,zhu2009predictable,zhu2011response,zhu2010selecting}.

\end{document}